\pdfoutput=1
\documentclass[11pt]{article}
\usepackage[]{acl}
\usepackage{times}
\usepackage{latexsym}
\usepackage{hyperref}
\usepackage[T1]{fontenc}
\usepackage[utf8]{inputenc}
\usepackage{microtype}
\usepackage{inconsolata}
\usepackage{graphicx}
\usepackage{amsmath}
\usepackage{amsfonts}
\usepackage[ruled,vlined]{algorithm2e}
\usepackage{enumitem}
\usepackage{multirow}
\usepackage{booktabs}
\usepackage{geometry}
\usepackage{bm}
\usepackage{xcolor}
\usepackage{colortbl}
\usepackage{tcolorbox}
\usepackage{color}
\definecolor{dt}{gray}{0.6}
\definecolor{lightgreen}{rgb}{0.88,1,0.88} 
\definecolor{lightred}{rgb}{1,0.94,0.94} 
\definecolor{lightgray}{gray}{0.9}

\usepackage{cleveref}
\crefformat{section}{\S#2#1#3}
\crefformat{subsection}{\S#2#1#3}
\crefformat{subsubsection}{\S#2#1#3}
\crefrangeformat{section}{\S\S#3#1#4 to~#5#2#6}
\crefmultiformat{section}{\S\S#2#1#3}{ and~#2#1#3}{, #2#1#3}{ and~#2#1#3}
\crefmultiformat{subsection}{\S\S#2#1#3}{ and~#2#1#3}{, #2#1#3}{ and~#2#1#3}
\Crefformat{figure}{#2Fig.~#1#3}
\Crefmultiformat{figure}{Figs.~#2#1#3}{ and~#2#1#3}{, #2#1#3}{ and~#2#1#3}
\Crefformat{table}{#2Tab.~#1#3}
\Crefmultiformat{table}{Tabs.~#2#1#3}{ and~#2#1#3}{, #2#1#3}{ and~#2#1#3}
\Crefformat{appendix}{Appx.~\S#2#1#3}
\crefmultiformat{appendix}{Appx.~\S#2#1#3}{ and~#2#1#3}{, #2#1#3}{ and~#2#1#3}
\crefformat{algorithm}{Alg.~#2#1#3}
\Crefformat{equation}{Eq.~#2#1#3}

\newcommand{\myparagraph}[1]{\vspace{0.3em}\noindent{{\bf #1.}}}
\newcommand{\modelname}{\textsc{DiverseEvol}\xspace}

\title{Self-Evolved Diverse Data Sampling for Efficient Instruction Tuning}

\author{
Shengguang Wu$^{1}$$^{2}$\thanks{$^*$Work done during internship at Alibaba Group.}, 
Keming Lu$^{1}$,
Benfeng Xu$^{1}$$^{3}$$^\ast$,
Junyang Lin$^{1}$,
Qi Su$^{2}$,
Chang Zhou$^{1}$
\\
$^1$Alibaba Group, 
$^2$Peking University,\\
$^3$University of Science and Technology of China 
\\
\texttt{\{wushengguang.wsg,lukeming.lkm,xubenfeng.xbf,junyang.ljy\}@alibaba-inc.com}, \\
\texttt{sukia@pku.edu.cn},
\texttt{ericzhou.zc@alibaba-inc.com} \\
}

\begin{document}
\maketitle

\begin{abstract}

Enhancing the instruction-following ability of Large Language Models~(LLMs) primarily demands substantial instruction-tuning datasets. However, the sheer volume of these imposes a considerable computational burden and annotation cost.
To investigate a label-efficient instruction tuning method that allows the model itself to actively sample subsets that are equally or even more effective,
we introduce a self-evolving mechanism \modelname. In this process, a model iteratively augments its training subset to refine its own performance, without requiring any intervention from humans or more advanced LLMs. 
The key to our data sampling technique lies in the enhancement of diversity in the chosen subsets, as the model selects new data points most distinct from any existing ones according to its current embedding space.
Extensive experiments across three datasets and benchmarks demonstrate the effectiveness of \modelname. Our models, trained on less than 8\% of the original dataset, maintain or improve performance compared with finetuning on full data. 
We also provide empirical evidence to analyze the importance of diversity in instruction data and the iterative scheme as opposed to one-time sampling. Our code is publicly available at \url{https://github.com/OFA-Sys/DiverseEvol.git}.


\end{abstract}

\section{Introduction} \label{sec:introduction}
Large Language Models~(LLMs) have demonstrated prowess in producing human-aligned  response to varied instructions. A pivotal technique for enhancing the instruction-following capabilities of LLMs is Instruction Tuning, which aligns the model with human preferences using data in the form of instruction-response pairs.

While massive instruction-tuning datasets exist, their vast quantity poses a significant computational burden, and their curation is itself a formidable challenge, given the meticulous labor involved in annotations.
Recent works shed light on data distillation, achieving similar or even better alignment performance relying on fewer instruction data, by mining compact subsets from extensive instruction datasets~\cite{zhou2023lima, cao2023instruction, chen2023alpagasus}. 
However, these works demand tremendous supervision from humans or advanced LLMs, such as GPT4~\citep{openai2023gpt4}, for selecting the ideal subset.

In contrast, our work introduces \modelname, a novel method featuring a \textbf{self-evolving} mechanism. In parallel to the approach in \citet{li2023self}, \modelname employs an iterative strategy, where the model relies on its current embedding space to augment its own training data samples that lead to an improved model in the next step. As such, instead of seeking external oversight, \modelname facilitates the model's \textbf{self-evolution}, as it actively selects data to refine its own performance through iterations.

Central to \modelname's design of data selection is the maintenance of high diversity. When curating a subset from a vast dataset, the key challenge is to ensure that this subset is as representative as possible. This indicates that data points within the subset must be diverse in order to ensure comprehensive coverage and simulate the effect of the entire dataset. 
Therefore, \modelname adopts a \textit{K}-Center-based~\citep{sener2017active} strategy that chooses data points characterized by the highest distance from any existing labeled data.

Our experiments span three distinguished instruction-tuning datasets curated by both human-annotation~\citep{DatabricksBlog2023DollyV2}, and Self-Instruct~\citep{alpaca, peng2023instruction}. Consistently, through \modelname, our models, trained on less than 8\% of the original datasets, match or outperform baselines trained on the entirety of the source datasets across all benchmarks.

Furthermore, our investigation yields two crucial findings. 
First, training dataset diversity is paramount for the success of instruction tuning. Our method's emphasis on diversity, quantified via the Vendi Score~\citep{friedman2022vendi}, correlates with enhanced model performance. 
Second, an iterative, evolving data sampling strategy outperforms direct, one-shot sampling. This evolution-driven approach, characterized by progressive data selection based on the model's current state, offers superior training outcomes.

In sum, our main contributions are three-fold:
\begin{itemize}[leftmargin=1em]
    \setlength\itemsep{-0.5em}
    \item A self-evolving, efficient data sampling pipeline, \modelname that requires significantly less data yet matches or surpasses the performance of models trained on complete datasets.
    \item A quantified demonstration of the essential role of dataset diversity in instruction-tuning, emphasizing the link between training data diversity and model performance.
    \item A revelation that iterative, evolving sampling outperforms static, one-time sampling, underscoring the advantages of progressive data selection for model improvement.
\end{itemize}

\section{Related Works} \label{sec:related_work}
\myparagraph{Instruction Tuning and Its Efficiency}
Instruction tuning is paramount for boosting the instruction-following capabilities of LLMs, and a range of methods have been utilized to curate large-scale datasets, extending from human annotations~\citep{DatabricksBlog2023DollyV2, kopf2023openassistant} to distillations from parent LLMs, such as Text-Davinci-003~\citep{alpaca}, GPT-3.5-TURBO~\citep{xu2023expertprompting}, and GPT4~\citep{peng2023instruction}. The Vicuna dataset~\citep{vicuna2023}, originating from ShareGPT's real-world interactions, serves as another exemplar in this regard. 
As the field advances, there's a growing inclination toward refining instruction tuning methods for better efficiency. 
\citet{alshikh2023becoming} shows that the instruction-tone is learned rather early without the need of training on full-sized dataset. \citet{zhou2023lima} yields promising results with only 1,000 manually curated instruction data. Concurrently, leveraging advanced LLMs for instruction data labeling has emerged as a trend, with endeavors like \citet{chen2023alpagasus} using ChatGPT for data rating and filtration, and others like \citet{lu2023instag} exploring diverse sampling based on open-world tag annotations. 
However, \modelname conducts diverse sampling with only its own supervision by a self-evolving mechanism while above methods necessitate external supervision from either humans and more advanced LLMs.


\myparagraph{Data Sampling Strategies}
Our work also draws inspirations from data-centric AI principles, emphasizing self-automated sampling strategies. These methodologies largely fall into two categories:
(1)~\textit{Uncertainty}-based approaches that prioritize datapoints the model's prediction deems ambiguous. Measures of the predictive uncertainty include maximum entropy (Entropy-Sampling,~\citealp{shannon2001mathematical}), lowest logits (Least-Confidence,~\citealp{wang2014new}), and minimal differences in the likelihood of top two probable labels (Margin-Sampling,~\citealp{netzer2011reading}).
(2)~\textit{Diversity}-based approaches that focus on a representative subset within the model's embedding space. Such strategies like \textit{K}-Center-Sampling~\citep{sener2017active} and Cluster-Margin~\citep{citovsky2021batch} have gained prominence.
In this work, we actively experiment above sampling strategies and empirically show that diversity-based sampling benefits the reduction of instruction data the most without harming model performance.

\begin{figure*}[htbp]
\centering
\includegraphics[scale=0.465]{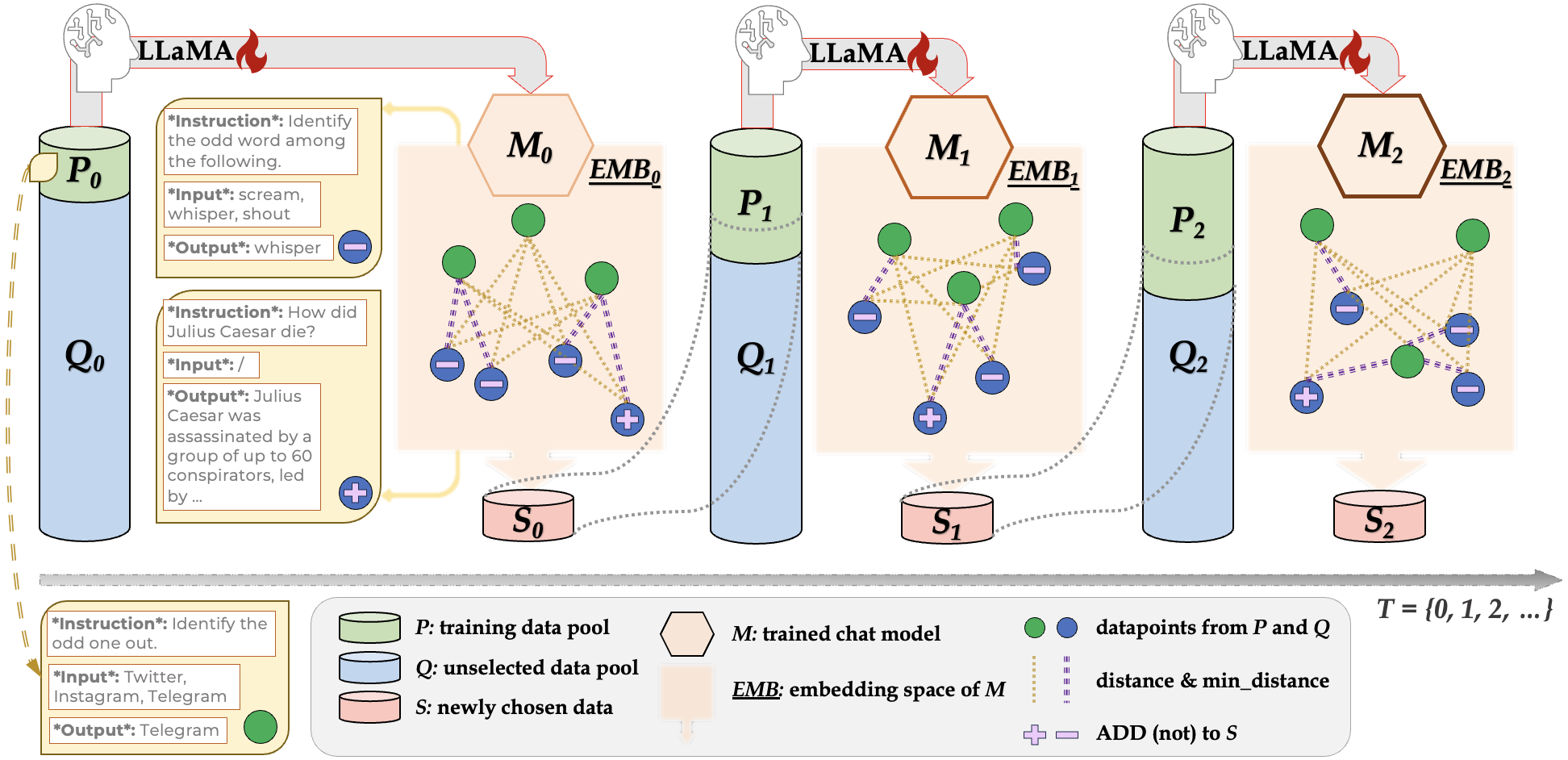}
\caption{Overview of our iterative \modelname: Starting with an initial training data pool $P_0$ and the remaining data $Q_0$ from the source dataset, we train a chat model $M_0$ and project all datapoints into its embedding space \underline{$EMB_0$}. Leverage \textit{K}-Center based selection~\Cref{sec:k_center_sampling} in this embedding space, a new set of datapoints $S_0$ is chosen from $Q_0$ and added to the next training data pool $P_1$ to instrution-tune the next chat model $M_1$. This process is repeated for $T$ steps, producing progressively augmented training data pool based solely on the model itself, which is then used to improve a more refined model with improved capabilities.}
\label{fig:model}
\end{figure*}

\section{\modelname} \label{sec:method}
In this section, we introduce \modelname, a self-evolved diverse sampling method for the +selection of instruction data. We first introduce instruction data selection as an iterative process~(\Cref{sec:overview}). Then, we lay out details about our \textit{K}-Center-based algorithm for the selection of training data~(\Cref{sec:k_center_sampling}). The overall workflow is illustrated in~\Cref{fig:model}.

\subsection{Iterative Instruction Data Selection} \label{sec:overview}
Our objective is to formalize instruction data mining as an iterative process, extracting from a vast source instruction dataset progressively according to a strategy. Given a collection of instruction-response pairs, denoted as $\mathcal{Z} = \{(x_i, y_i)\}_{i \in \mathbb{N}}$, where each $(x_i, y_i)$ represents a specific instruction-response pair, we define $\mathbb{N} = \{1, \ldots, n\}$ as the size of the initial source instruction dataset. The iterative procedure revolves around two data containers: the training data pool $P_t$ up to iteration step $t$ and the container of unselected data points, $Q_t$. 
At each iteration $t$, a selection function (i.e., strategy) $A$ determines which data points, $\mathcal{S} = \{s_j\}_{j \in \mathbb{K}}$, with $\mathbb{K} = \{1, \ldots, k\}$, are integrated into the training data pool $P_{t+1}$ for the next step. This expanded pool then serves as the training set for the next model iteration, $M_{t+1}$.

Beginning with a randomized data pool, $P_0$, to train the initial model $M_0$, every subsequent step employs model $M_t$, the current training pool $P_t$, and the comprehensive dataset $\mathcal{Z}$ to inform function $A$, which then outputs new data points $S_t$ to be added to the training pool for the next iteration $P_{t+1}$, as in: $S_{t} = A(\mathcal{Z}, P_t, M_t) ;  P_{t+1} = P_t \cup S_t$. 
Thus, each iteration consists of two operations:
1. Deduce new data points $S_{t}$ to merge into $P_{t+1}$, informed by the previously trained model $M_{t}$.
2. Train the subsequent chat model, $M_{t+1}$, with the updated data pool $P_{t+1}$.

The efficacy of this approach hinges on the selection function $A$ that determines the additional $k$ data points for each training iteration. As $P$ grows both in volume and, crucially, in diversity (as stressed by our method, see~\Cref{sec:k_center_sampling}), the resulting chat model continuously refines its capabilities.

\subsection{Selection Algorithm: \textit{K}-Center-Sampling} \label{sec:k_center_sampling}
Central to \modelname is our selection function $A$ based on the \textit{K}-Center-Sampling method ~\citep{sener2017active}, as detailed in \Cref{algo:k_center}. The selected subset must aptly represent the broader dataset to ensure that models trained on reduced subsets rival those trained on the complete dataset. Thus, our function $A$ strives to amass a highly diverse subset of the source dataset, reminiscent of the facility location problem~\citep{wolf2011facility, wei2013using}.

With a given set of training data points, $P_t$, function $A$ identifies novel data points $S_t$ that, when combined with $P_t$, provide a representative sample of the source dataset. This entails selecting newly added data that is as \textbf{different} as possible from any of the existing data points. The "difference" from existing data points is quantified by the closest distance of a candidate datapoint (i.e., an as-yet unchosen data point from $Q_t$) to any existing training data in $P_t$. In other words: the distance to its nearest neighboring datapoint $P_t$. 
Therefore, our objective for $A$ at iteration $t$ can be succinctly articulated as:

\textbf{Objective:} \textit{From a candidate pool, choose $k$ data points in such a way that the distances to their respective nearest existing training data points are maximized.}

\begin{equation}
\max \sum_{1 \leq i \leq k} \min _{j \in P_t} \Delta\left(\mathbf{s}_{i}, \mathbf{p}_{j}\right)
\end{equation}

Our function aims to designate each of the $k$ new data points as a unique center within the full training pool. Consequently, it seeks to maximize the minimum distance from each new data point in $S_t$ to any existing training data point in $P_t$. As formulated below, for $k$ data points to be selected from the candidate datapoint pool $Q_t$, we select:

\begin{equation}
\mathop{\arg\max}_{i \in Q_t} \min_{j \in P_t} \Delta\left(\mathbf{s}_{i}, \mathbf{p}_{j}\right)
\end{equation}

The embeddings produced by the currently trained model $M_t$ guide our selection since the distance between samples, denoted as $\Delta$, is computed based on the output hidden states of $M_t$ after average pooling over all token positions, which provides a more suitable embedding space for existing data. As such, data points added to the training set ensure to best supplement the existing dataset according to the model's current understanding. This iterative procedure facilitates the model's \textbf{evolution}, as it incorporates insights from prior iterations to refine its performance.


\begin{algorithm}
  \caption{Iterative \textit{K}-Center-Sampling for \( T \) Steps}
  \KwIn{$Z$: entire source dataset; $M_{pretrain}$: foundation LLM; $k$: budget for new data points; $T$: total number of iterations}
  \KwOut{\textit{Series} \( P = \{P_0, P_1, \ldots, P_T\} \); \textit{Series} \( M = \{M_0, M_1, \ldots, M_T\} \)}
  \textbf{Initialize:} $P_0$: \( k \) data points randomly sampled from \( Z \); $Q_0 = Z \setminus P_0$ \\
  \For{\( t = 0 \) \KwTo \( T-1 \)}{
    \textbf{Finetune:} \( M_{pretrain} \) using \( P_t \) to get \( M_t \)\\
    \textbf{Select data points:} \\
    \Indp 
        \textbf{initialize:} $S_t = \emptyset$; $Q'_t = Q_t$ \\
        \Repeat{$|S_t| = k$}{
            $s = \mathop{\arg\max}_{i \in Q'_t} \min_{j \in P_t} \Delta\left(\mathbf{s}_{i}, \mathbf{p}_{j}\right)$ \\
            $S_t = S_t \cup \{s\}$ \\
            $Q'_t = Q'_t \setminus \{s\}$ \\
        }
    \Indm
    \textbf{Update Pools:} \\
    \Indp 
        $P_{t+1} = P_t \cup S_t$ \\
        $Q_{t+1} = Z \setminus P_{t+1}$ \\
    \Indm
  }
  \KwRet{Series \( P \), Series \( M \)}
  \label{algo:k_center}
\end{algorithm}

\section{Experiments} \label{sec:experiments}
\begin{table*}[htbp]
\centering
\footnotesize
\setlength{\tabcolsep}{7pt}
\begin{tabular}{lccccccccc}
\toprule
\textbf{Sampling Strategy} & \multicolumn{3}{c}{\textbf{Vicuna-Bench}} & \multicolumn{3}{c}{\textbf{Koala-Bench}} & \multicolumn{3}{c}{\textbf{Wizardlm-Bench}} \\
\cmidrule(lr){2-4} \cmidrule(lr){5-7} \cmidrule(lr){8-10}
& \textit{\textbf{RS}} & \textit{\textbf{WTR}}& \bm{$N_{best}$} & \textit{\textbf{RS}} & \textit{\textbf{WTR}}& \bm{$N_{best}$} & \textit{\textbf{RS}} & \textit{\textbf{WTR}}& \bm{$N_{best}$} \\
\midrule
\multicolumn{10}{c}{\textit{Source Dataset = Databricks-Dolly-15K}} \\
\midrule
\rowcolor{lightgray}
\texttt{*Full Data} & \underline{73.84}& 5.00& $15011$ & \underline{57.90}& 3.33& $15011$ & \underline{58.73}& 3.21& $15011$ \\
Random & 73.06 & \underline{6.25\#}& $700$& 53.11 & 3.33*& $900$& 56.02& \underline{4.59*}& $1100$\\
Least-Confidence & 46.68 & 0.00& $100$& 36.01 & 2.27*& $1100$& 40.08 & 1.38& $800$\\
 Margin-Sampling& 69.67& 3.75& $400$& 52.29& \underline{5.00}& $600$& 53.53& 3.21*&$900$\\
\textbf{\textit{K}-Center (\modelname)} & \textbf{79.69}& \textbf{20.00}& $700$& \textbf{62.29} & \textbf{6.67}& $1100$& \textbf{62.94} & \textbf{8.26}& $700$\\
\midrule
\multicolumn{10}{c}{\textit{Source Dataset = SelfInstruct-Davinci-52K}} \\
\midrule
\rowcolor{lightgray}
\texttt{*Full Data} & 73.03 & \underline{2.50}& $52002$ & \textbf{69.50}& 3.89& $52002$ & \underline{61.59}& 5.05& $52002$ \\
Random & \underline{75.43}& \textbf{7.50*}& $800$& 62.33 & \underline{5.56}& $900$& 58.60 & \underline{5.96*}& $500$\\
Least-Confidence & 64.27 & \underline{2.50}& $600$& 43.27 & 3.33\#& $100$& 49.26 & 5.05*& $500$\\
 Margin-Sampling& 68.98& \underline{2.50*}& $1000$& 55.22& 2.78& $1000$& 53.98& 2.75&$1000$\\
\textbf{\textit{K}-Center (\modelname)} & \textbf{79.16}& \textbf{7.50*}& $1000$& \underline{66.95}& \textbf{6.11*}& $1100$& \textbf{63.08} & \textbf{7.80*}& $700$\\
\midrule
\multicolumn{10}{c}{\textit{Source Dataset = SelfInstruct-GPT4-52K}} \\
\midrule
\rowcolor{lightgray}
\texttt{*Full Data} & \underline{90.28}& 46.25& $52002$ & \textbf{80.33}& 10.56& $52002$ & \textbf{75.00}& 12.84& $52002$ \\
Random & 90.21 & \underline{48.75\#}& $500$& 77.31 & \underline{12.78}& $800$& 71.95 & \textbf{14.68*}& 1000 \\
Least-Confidence & 79.11 & 17.5*& $1100$& 55.57 & 4.44\#& $800$& 58.33 & 6.88& $100$\\
 Margin-Sampling& 82.43& 33.75\#& $600$& 63.10& 7.22& $1000$& 65.01& 8.26&$1000$\\
\textbf{\textit{K}-Center (\modelname)} & \textbf{91.69} & \textbf{50.00\#}& $400$& \underline{79.01}& \textbf{14.44*}& $1100$& \underline{73.36}& \underline{13.76}& $1000$\\
\bottomrule
\end{tabular}
\caption{Comparison of the \textit{K}-Center-based \modelname method with alternative sampling strategies and "strong" baselines using the full source data. Metrics include relative scores (\textit{RS}), win-and-tie rate (\textit{WTR}), and optimal data sizes (\bm{$N_{best}$}) behind the peak \textit{RS}. If the best \textit{WTR} is obtained with fewer data than \bm{$N_{best}$}, it is marked with *, otherwise \#. The gray-shaded rows are models using the entire source datasets as strong benchmarks. The best results are in \textbf{bold}; the second-best is \underline{underlined}. Our \modelname approach consistently delivers high-quality results, matching or surpassing the strong baselines, with substantially fewer training samples.}
\label{tab:main_results_wtr}
\end{table*}

\begin{figure*}[t]
\centering
\includegraphics[width=\linewidth]{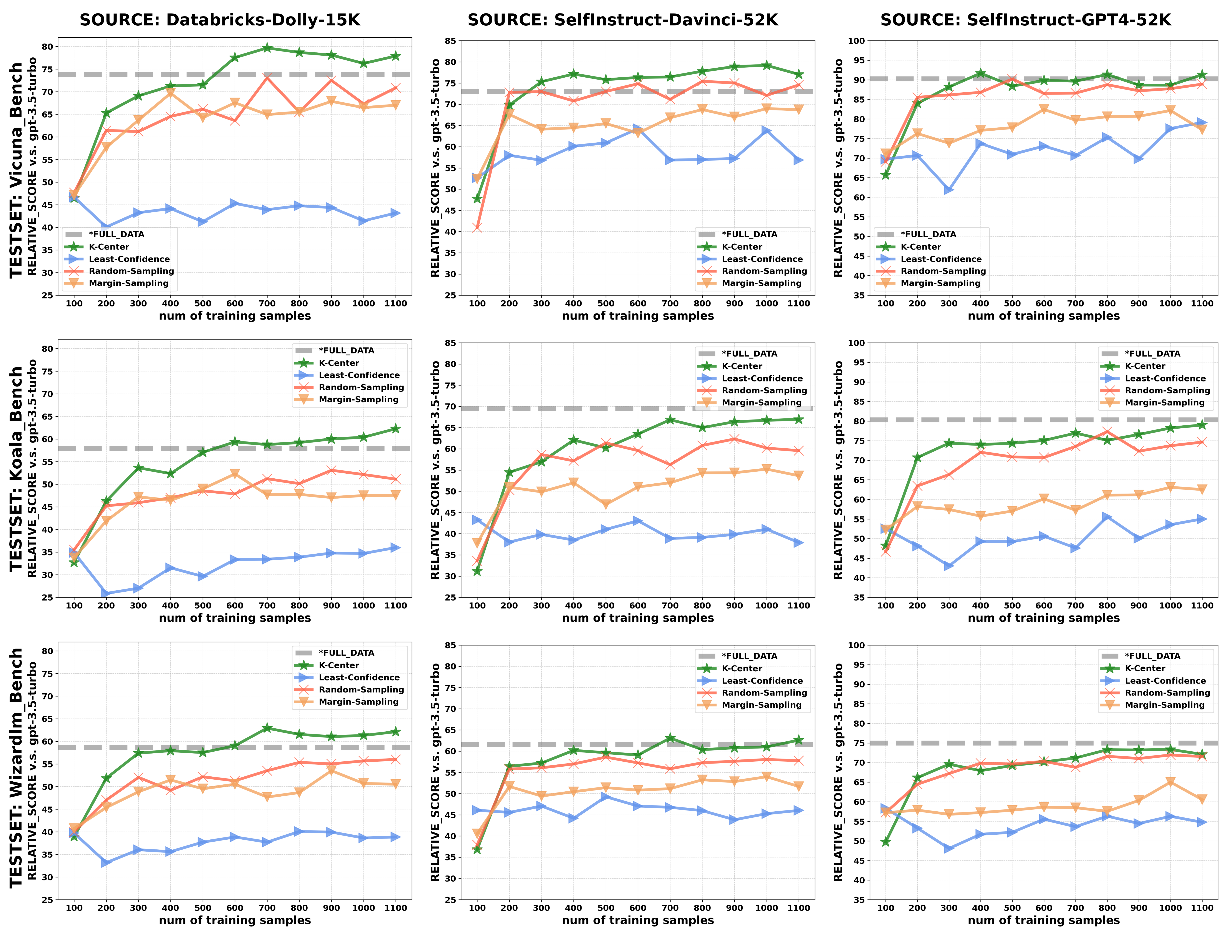}
\caption{Performance evolution of chat models across various source datasets using our proposed \textit{K}-Center based \modelname and alternative sampling approaches. The Y-axis represents relative scores (\textit{RS}) with respect to ChatGPT, while the X-axis indicates the number of training samples. The curves demonstrate the rapid proficiency gains achieved by the \modelname approach, matching or often outpacing strong baselines (\texttt{*Full Data}) trained on the full dataset with only a significantly small fraction of the data.}
\label{fig:evolution_rs}
\end{figure*}

In this section, we introduce the experimental setup~(\Cref{sec:experimental_setup}), main results~(\Cref{sec:evoluation_of_model_performance}), and conduct rich analyses about the effectiveness of \modelname that can be attributed to its central designs of data diversity and iterative sampling~(\Cref{sec:diversity_and_iterativeness}).

\subsection{Experimental Setup} \label{sec:experimental_setup}

\myparagraph{Datasets}
Three prominent open-source instruction-tuning datasets serve to validate the effectiveness of \modelname. These include both human-annotated data (Databricks-Dolly,~\citealp{DatabricksBlog2023DollyV2}) and machine-generated (SelfInstruct-Davinci,~\citealp{alpaca}, SelfInstruct-GPT4,~\citealp{peng2023instruction}). Statistics are detailed in \Cref{tab:data-stats}.

\begin{table}[htbp]
\centering
\footnotesize
\begin{tabular}{lcc}
\toprule
\textbf{Source Datasets} & \textbf{\# Samples} & \textbf{Annotator/Engine} \\
\midrule
Databricks-Dolly & 15011 & human \\
SelfInstruct-Davinci & 52002 & Text-Davinci-003 \\
SelfInstruct-GPT4 & 52002 & GPT-4 \\
\bottomrule
\end{tabular}
\caption{Source datasets used in our experiments.}
\label{tab:data-stats}
\end{table}

\myparagraph{Baselines}
As a data sampling method, we introduce strong baselines that correspond to chat models directly trained on the full-sized source datasets, including LLaMA-7B finetuned on Databricks-Dolly, SelfInstruct-Davinci, and SelfInstruct-GPT4 respectively.
For comparison, our \textit{K}-Center-based method, which prioritizes diversity, is also benchmarked against the following:
(1)~Random-Sampling: stochastically selects data points at each iteration.
(2)~Least-Confidence~\citep{culotta2005reducing}: samples data points the current model exhibits least confidence in, measured by the average max-logit value across the predicted token sequence.
(3)~Margin-Sampling~\citep{netzer2011reading}: chooses data points whose logits obtained by current model show minimal differences in the likelihood of top two probable tokens.

\myparagraph{Benchmarks} 
We test our method on three distinct benchmarks: Vicuna-Bench~\citep{vicuna2023}, Koala-Bench~\citep{geng2023koala}, and Wizardlm-Bench~\citep{xu2023wizardlm} to ensure a extensive evaluation and help minimize test set biases. Alongside these, we adopt an evaluation framework, as in prior works~\citep{vicuna2023,alpacafarm,mt_bench,xu2023expertprompting}, with GPT4-Judge ($J$) scoring two model responses (template detailed in Appendix~\ref{sec:gpt4-judge-template}). We also randomly permute the order of the two answers to counteract potential position biases in GPT4’s judgement. Specifically, we compare the answers of all chat models ($A^{\text{model}}$) to those generated by GPT3.5-TURBO ($A^{\text{chatgpt}}$), a general competitor. We then compute Relative Score (RS) and Win-And-Tie-Rate (WTR) vs. ChatGPT as metrics to assess instruction-following capabilities. 

\begin{itemize}[leftmargin=1em]
    \setlength\itemsep{-0.5em}
    \item \textbf{Relative Score~(RS)} vs. ChatGPT: Compares the chat model's performance with ChatGPT based on their scores, formulated as:
    \begin{equation}
    \textrm{RS} = \frac{\sum_{q \in \text{testset}} J(A^{\text{model}}_q)}{\sum_{q \in \text{testset}} J(A^{\text{chatgpt}}_q)}
    \end{equation}
    \item \textbf{Win-And-Tie Rate~(WTR)} vs. ChatGPT: Measures the frequency at which the chat model outperforms (WIN) or matches (TIE) the performance of ChatGPT:
    \begin{equation}
    \textrm{WTR} = \frac{\sum_{q \in \text{testset}} \mathbb{I}(J(A^{\text{model}}_q) \geq J(A^{\text{chatgpt}}_q))}{|\text{testset}|}
\end{equation}
\end{itemize}

\myparagraph{Configurations} 
All our experiments utilize LLaMA-7B~\citep{touvron2023llama} as the foundation LLM ($M_{pretrain}$). Unless stated otherwise, all iterative data sampling begins with an initial pool $P_0$ of $100$ random samples. It spans $T = 10$ iterations with a new data point budget $k = 100$. For instruction-tuning each chat model, we finetune the LLaMA model for 3 epochs with the batch size set to 128 and the learning rate set to $2\times10^{-5}$. The Alpaca-style template~\citep{alpaca} is adopted to prepare input from the instruction data.

\subsection{Main Results} 
\label{sec:evoluation_of_model_performance}

Utilizing our \modelname approach, chat models evolve in their instruction-following capability as the training data pool progressively augments through our \textit{K}-Center-Sampling strategy. 

\Cref{tab:main_results_wtr} compares our \textit{K}-Center-based \modelname method with alternative sampling strategies and strong baselines trained on full source data (\texttt{*Full Data}). The metrics reported include Relative Scores (\textit{RS}), Win-and-Tie Rates (\textit{WTR}), and the optimal data sizes (\bm{$N_{best}$}) associated with peak \textit{RS}. With the \textit{K}-Center-based \modelname strategy, our chat models frequently match or exceed the performance of the strong baselines with far fewer training samples. 
On the human-annotated source dataset \textit{Databricks-Dolly-15K}, our method consistently achieves the best \textit{RS} and \textit{WTR} across benchmarks, surpassing the baseline finetuned on the entire 15K data by a considerable margin with merely 700 or 1100 samples, corresponding to less than 8\% data size.
On the \textit{SelfInstruct-52K} data generated by \textit{Text-Davinci-003} or \textit{GPT4}, \modelname achieves similar effects of top performance surpassing the strong baselines on the majority of metrics using only 2\% or less of the 52K source data ($\leq$ 1100 samples). 
Even on benchmarks where our method does not stand out as the best performer, it achieves at least the second-best results behind the strong baselines by a small margin, such as in the case of \textit{RS} with the highest gap of mere 2.55 on Koala-Bench using the \textit{SelfInstruct-Davinci} source data. This unambiguously shows the effectiveness and efficiency of our proposed \modelname data selection strategy. 
In contrast, other sampling strategies like random sampling or confidence-based selection (e.g., Least-Confidence, Margin-Sampling as discussed in~\Cref{sec:experimental_setup}) tend to underperform or at best only seldom match the strong baselines, which largely falls behind \modelname's overall performance. 

\Cref{fig:evolution_rs} provides a complementary view to \Cref{tab:main_results_wtr}, illustrating the exact trajectory of performance evolution (measured by \textit{RS}) with iteratively extended training data pool. The trend line in this figure is revealing. 
Our \textit{K}-Center based \modelname models (marked in green) start to match or surpass the strong baselines trained on the complete dataset (\texttt{*Full Data}) remarkably quickly, namely in only a few iterative steps, requiring several hundred samples selected from the source dataset. 
On the source dataset \textit{Databricks-Dolly-15K}, our method manages to match the upper bound-baseline with only 600 samples (4\%) across test sets. 
Compared with alternative sampling strategies, our \textit{K}-Center-based \modelname method also consistently stands out as the top-performing curve, showing better scores throughout the iteration, regardless of source datasets or testing benchmarks.

 \begin{figure*}[t]
\centering
\includegraphics[width=\linewidth]{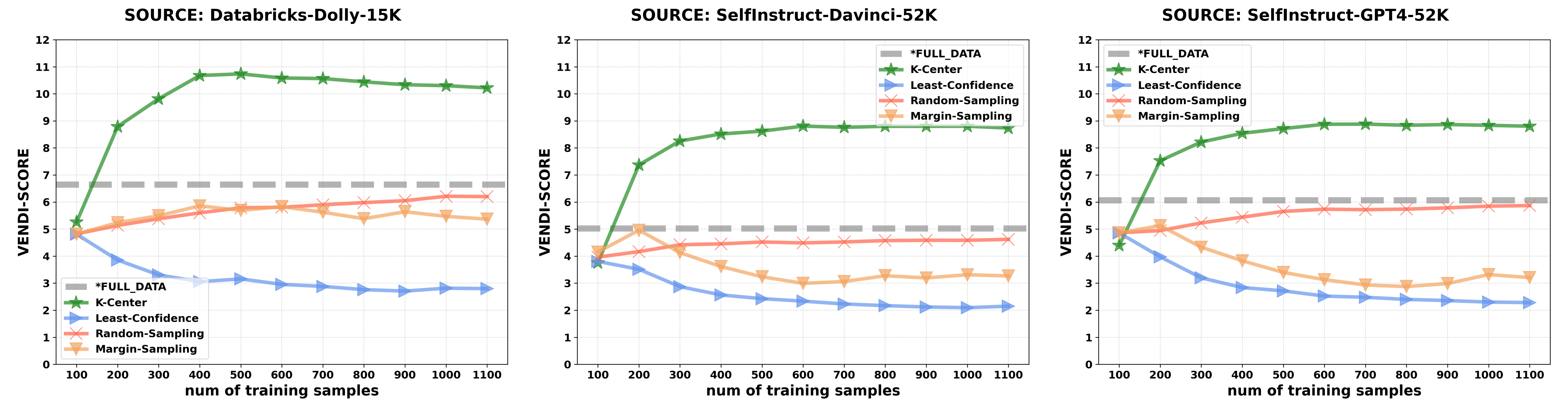}
\caption{Diversity evolution in the selected training data pool from three source datasets. The Y-axis denotes the Vendi-Score for measuring diversity, and the X-axis shows increasing data size. The gray line (\texttt{*Full Data}) represents original source dataset diversity. The contrasting curves highlight our \textit{K}-center approach's early and sustained enhancement of data diversity.}
\label{fig:evolution_vendi}
\end{figure*}

\begin{table*}[t]
\centering
\footnotesize
\begin{tabular}{llccccccccc}
\toprule
\textbf{\textit{K}-Center} & & \multicolumn{3}{c}{\textbf{Vicuna-Bench}} & \multicolumn{3}{c}{\textbf{Koala-Bench}} & \multicolumn{3}{c}{\textbf{Wizardlm-Bench}} \\
\cmidrule(lr){3-5} \cmidrule(lr){6-8} \cmidrule(lr){9-11}
& \multirow{1}{*}{\bm{$N$}} & 300 & 700 & 1100 & 300 & 700 & 1100 & 300 & 700 & 1100 \\ 
\midrule
\textbf{Iterative (\modelname)} & \bm{$RS$} & \textbf{69.09}& \textbf{79.69}& \textbf{77.90}& \textbf{53.65}& \textbf{58.78}& \textbf{62.29}& \textbf{57.42}& \textbf{62.94}& \textbf{62.15}\\
\textbf{One-Time Direct Sampling} & \bm{$RS$} & 67.38& 73.90& 73.21& 51.42& 58.10& 57.56& 50.94& 61.82& 60.97\\
\bottomrule
\end{tabular}
\caption{Comparison of performance between the dynamic, iterative sampling scheme as in \modelname and one-time data selection method of directly sampling to a given data size. With the same \textit{K}-Center selection algorithm, this table shows that the iterative approach consistently outperforms the method of direct sampling for once across different data volumes, highlighting the importance of iterative feedback in improving chat model capabilities.}
\label{tab:direct_vs_iterative}
\end{table*}

\begin{figure*}[t]
\centering
\includegraphics[scale=0.4]{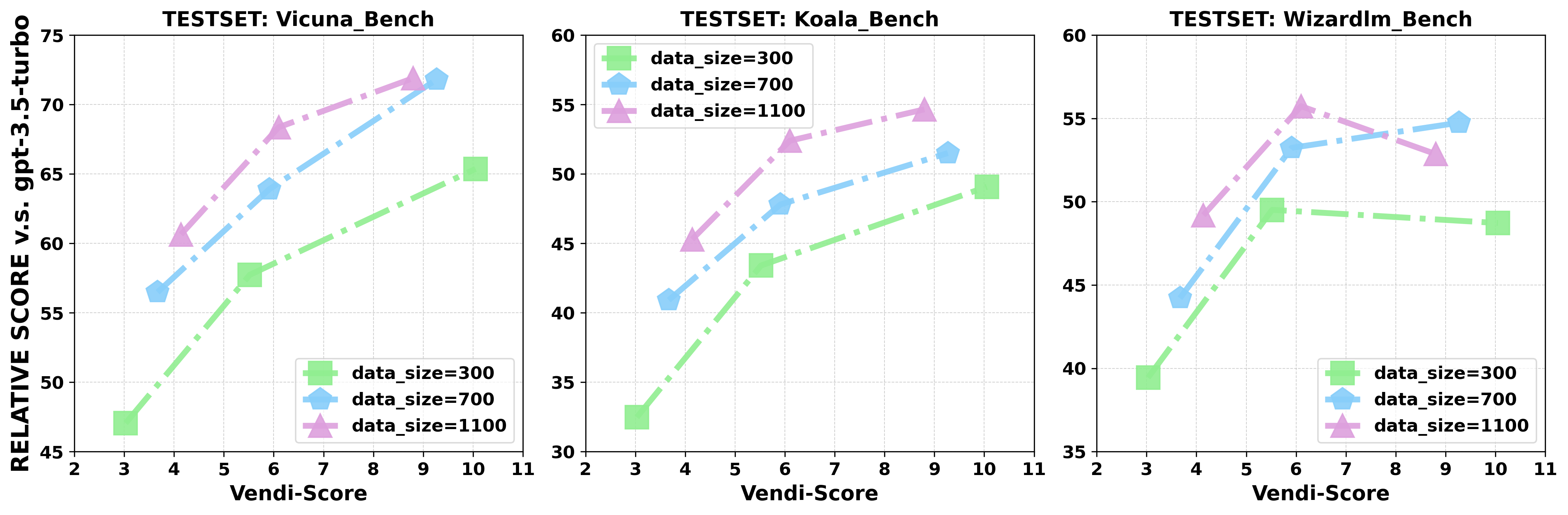}
\caption{Performance of instruction-tuned chat models in relation to Vendi-Score of their training datasets, illustrating the influence of data diversity. The three distinct curves correspond to training data volumes of $300$, $700$, and $1100$. A consistent trend of performance enhancement is observed with increased dataset diversity across most benchmarks, with only minor deviations seen on the Wizardlm-Bench.}
\label{fig:vendi_score_and_performance}
\end{figure*}

\subsection{Analyses} 
\label{sec:diversity_and_iterativeness}

We provide further analyses of the two main factors behind the effectiveness of \modelname, namely: diversity of selected datasets, and the dynamic iteration scheme.

\myparagraph{Diversity}
Based on the main results reported in \Cref{tab:main_results_wtr} and \Cref{fig:evolution_rs}, we believe that maintaining high diversity in the training data pool is crucial for a successful instruction-tuning dataset. This is also exactly the design principle behind our \textit{K}-Center based \modelname that seeks to find the most representative subset of a source data pool, constituting the most diverse cover of the source dataset (\Cref{sec:k_center_sampling}). Given that diversity is a focal point in our method, we also explicitly assess data diversity using an automatic metric, \textbf{Vendi-Score}~\citep{friedman2022vendi} that measures the datapoint distribution's diversity based on their embeddings' similarity matrix. 
To testify to the pivotal role of diversity, we thus conduct empirical analyses from the following two angles.

First, we use the above diversity metric to quantitatively measure the level of data diversity achieved by our \textit{K}-Center-based method, compared to the original dataset diversity and other sampling methods.
In \Cref{fig:evolution_vendi}, we present the Vendi-Score of the maintained training data pool $P_t$ at each iteration step $t$, in line with the X-axis in \Cref{fig:evolution_rs}. As shown in the figure, our \textit{K}-Center data selection algorithm~(\Cref{algo:k_center}) significantly boosts the diversity of the training data pool at an early stage, surpassing the diversity of the original source dataset and all other sampling methods. This demonstrates the effectiveness of our \textit{K}-center-based sampling in selecting datapoints that constitute the most diverse cover of the source dataset.

Second, to further demonstrate the diversity of the training dataset as a key contributor to model performance, we directly control the Vendi-Score as a diversity variable and report how varying the level of diversity in the training dataset leads to varying instruction-tuned chat model performance. 
Using \textit{Databricks-Dolly} as an example source dataset, we perform independent random sampling, devoid of any algorithmic influence, for multiple iterations to achieve specific Vendi-Scores for predetermined training data sizes. Our experiment comprises three distinct training data volumes: $300$, $700$, $1100$. For each volume, we target three levels of diversity, measured by Vendi-Score of ranges: $[3,4]$, $[5,6]$, and $[9,10]$. A negligible deviation of $\pm0.2$ is observed, because larger data sizes make it harder to mine more or less diverse samples given the randomness of the procedure. Subsequently, we train chat models using datasets behind the highest, median, and lowest range of Vendi-Score, representing high, medium, and low data diversity, respectively. 
In \Cref{fig:vendi_score_and_performance}, we show the resulting chat model performance measured by Relative Score~(RS) v.s. ChatGPT in regard to Vendi-Score of its training dataset, signifying the level of diversity. Each curve represents a controlled total training data size. Evidently, the degree of diversity in the training data pool significantly influences the resulting chat model's performance regardless of data volume. We observe an nearly consistent boost of chat model performance as we maintain a more diverse training data pool almost across testing benchmarks, except for marginal deviations on the Wizardlm-Bench. The sheer elevation of \textit{RS} as a result of increased dataset diversity is striking, often reaching over 10 points, especially from the very lowest range of Vendi-Score to the medium level. This effectively proves data diversity as a key factor in boosting instruction-tuned chat model capability.

\myparagraph{Dynamic Iteration}
Another distinguishing aspect of our methodology is its iterative nature in data selection, which we demonstrate is crucial in bolstering the chat model's ability to follow instructions. 
Using the \textit{Databricks-Dolly} source dataset as an example case, we contrast our primary iterative approach, where the chat model's data pool incrementally expands, against an alternative strategy where data is directly sampled at three different volumes: $300$, $700$, and $1100$. Both methods employ the same \textit{K}-Center selection method, with the initial $100$ samples chosen randomly.

\Cref{tab:direct_vs_iterative} vividly demonstrates the differences in performance. Regardless of the final training data size, our proposed iterative approach (\modelname), mirroring the results in \Cref{tab:main_results_wtr} with corresponding $N_{best} = N$, consistently outperforms the method of directly sampling the same data volume (One-Time Sampling). Notably, while the \textit{K}-Center sampling technique remains identical across both approaches, the obvious performance variance underscores the pivotal role of iterative feedback. Such signals, derived from the trained chat model at every iterative step, guides subsequent data selections and establishes a progressive learning mechanism that capitalizes on insights from prior iterations. This contrasts sharply with direct sampling, which misses out on leveraging the experience accrued from past models, leading to suboptimal results. Therefore, our approach enables models to truly "evolve" itself over iterations, using insights from previous stages to inform future training data selection. This iterative feedback loop starkly outperforms a one-off decision-making process, underlining its essential role in enhancing model performance.

\section{Conclusion} \label{sec:conclusion}
We introduced \modelname, a self-evolving method for efficient instruction tuning of LLMs. Relying on an iterative scheme, \modelname progressively improves itself by selecting diverse subsets from vast instruction data using the \textit{K}-Center strategy without seeking any external supervision. Empirical results affirm that, with less than 8\% of the original data size, our method matches or surpasses strong baselines in performance.
Future endeavors can delve into leveraging our method on larger instruction datasets for potentially even more refined results. Building upon the foundation laid by \modelname, more advanced algorithms of diverse sampling also promise to enhance model performance further.

\section*{Limitations} \label{sec:limitations}
The \textit{K}-Center sampling method in \modelname involves computing distances between high-dimensional embeddings of datapoints. If the source dataset further increases in size, this computation may impose a considerable expense on the GPU memory. 
Furthermore, our evaluation outcomes rely heavily on GPT4-judge. Despite our attempts to obtain a more deterministic result by setting the querying temperature to 0, and to address position-bias through two-time querying with model responses in alternating positions, the evaluation process may still be influenced by inherent biases within the GPT4 model.


\section*{Ethics Statement}

All data, pretrained models, and results are collected and processed according to the respective data and API usage policy.
Finetuned models with \modelname may create toxic or unsafe contents. Therefore, outputs from these models need careful verification before being applied to real-world applications

\bibliography{main}

\appendix

\section{GPT4-Judge Template}
\label{sec:gpt4-judge-template}

We conduct automatic evaluation of chat model's performance using GPT4 as judge (\Cref{sec:experimental_setup}). Given a question (i.e., instruction) from test set and answers generated by two models, here's the template we used, adapted from~\citep{vicuna2023}:

\begin{tcolorbox}[colback=black!5!white,colframe=black!75!black,title=Template for GPT4-Judge]
\lbrack Question\rbrack\\
\texttt{{\color{red}\{\text{instruction}\}}}\\
\\
\lbrack The Start of Assistant 1's Answer\rbrack \\
\texttt{{\color{red}\{\text{answer-of-chatbot1}\}}} \\
\lbrack The End of Assistant 1's Answer\rbrack \\
\\
\lbrack The Start of Assistant 2's Answer\rbrack \\
\texttt{{\color{red}\{\text{answer-of-chatbot2}\}}} \\
\lbrack The End of Assistant 2's Answer\rbrack \\
\\
\lbrack System\rbrack \\
We would like to request your feedback on the performance of two AI assistants in response to the user question displayed above. Please rate the helpfulness, relevance, accuracy, level of details of their responses. Each assistant receives an overall score on a scale of 1 to 10, where a higher score indicates better overall performance. Please first output a single line containing only two values indicating the scores for Assistant 1 and 2, respectively. The two scores are separated by a space. In the subsequent line, please provide a comprehensive explanation of your evaluation, avoiding any potential bias and ensuring that the order in which the responses were presented does not affect your judgment.
\end{tcolorbox}

Throughout our experiments, the specific model versions of our OpenAI's API calls are: \textit{GPT-3.5-TURBO-0613} and \textit{GPT-4-0613}.


\end{document}